\documentclass[review,authoryear]{elsarticle}

\pdfoutput=1
\usepackage{amsmath}
\usepackage{amstext}
\usepackage{rotating}
\usepackage[usenames,dvipsnames]{pstricks}
\usepackage{epsfig}
\usepackage{pst-plot} 
\usepackage{xcolor}

\journal{Neural Networks}

\begin{document}

\begin{frontmatter}
\title{Short-Term Memory Through Persistent Activity: Evolution of Self-Stopping and Self-Sustaining Activity in Spiking Neural Networks}
\author[rvt]{Julien Hubert\corref{cor:jh}}
\ead{jhubert@sacral.c.u-tokyo.ac.jp}
\author[rvt]{Takashi Ikegami}
\ead{ikeg@sacral.c.u-tokyo.ac.jp}
\address[rvt]{The University of Tokyo, Graduate School of Arts And Science, Department of General Systems Studies, 3-8-1 Komaba, Meguro-ku, Tokyo, 153-8902, Japan}
\cortext[cor:jh]{Corresponding author}

\begin{abstract}
Memories in the brain are separated in two categories: short-term and long-term memories. Long-term memories remain for a lifetime, while short-term ones exist from a few milliseconds to a few minutes. Within short-term memory studies, there is debate about what neural structure could implement it. Indeed, mechanisms responsible for long-term memories appear inadequate for the task. Instead, it has been proposed that short-term memories could be sustained by the persistent activity of a group of neurons. In this work, we explore what topology could sustain short-term memories, not by designing a model from specific hypotheses, but through Darwinian evolution in order to obtain new insights into its implementation. We evolved 10 networks capable of retaining information for a fixed duration between 2 and 11s. Our main finding has been that the evolution naturally created two functional modules in the network: one which sustains the information containing primarily excitatory neurons, while the other, which is responsible for forgetting, was composed mainly of inhibitory neurons. This demonstrates how the balance between inhibition and excitation plays an important role in cognition.
\end{abstract}


\end{frontmatter}

\section{Introduction}
To achieve adaptability, the brain possesses different types of memory based on the kind of knowledge that must be acquired. These memory systems can be separated into two categories depending on how long they retain information: long-term or short-term. In long-term memory (LTM), we find declarative memories, such as episodic and semantic memories, and non-declarative memories such as procedural, priming and perceptual, or conditioning(\cite{squire04memorysystems}). In terms of shot-term memories (STM), we find sensory working memories, such as visual working memory or auditory working memory, and the non-sensory working memory also referred to as the scratch pad of the brain for its capacity to retain and process information for a limited amount of time (\cite{pasternak05sensory,baddeley12workingmem}). The goal of this research is to look into the neural mechanisms of short-term memory.
\par There are many theories concerning STM, and no clear consensus yet on which mechanisms are employed in the brain (\cite{durstewitz00workingmem, barbieri2008can}). The most likely situation is that the brain employes different mechanisms in different areas. There is at least one idea that seems shared by all theories: the mechanisms for LTM cannot implement STM because of the timescale they are working on. Indeed, the time required to store new items using synaptic based plastic mechanisms, such as spike timing dependent plasticity (STDP)(\cite{song00competitive}), shows an onset of the structural modifications of 2-3s, which will develop for 30s afterward  (\cite{hanse94onset}), while STM does not show any latency. As such, the main theoretical framework is focused on STM being implemented through persistent activity within a group of neurons \cite{wang2001synaptic, funahashi89mnemonic}. Other theories relying on fast synaptic changes (\cite{sugase2008short}), or on synaptic facilitation (\cite{mongillo2008synaptic}) have also been recently proposed.
\par In this paper we are interested in the persistent non-synaptic theories of STM. Currently, modeling is used to test specific hypotheses on how STM could be implemented in the brain. Our work aims at taking the opposite approach to what is generally found in computational neurosciences. Rather than proposing a new hypothesis, we will evolve a neural network for a STM task using a genetic algorithm(GA)(\cite{holland75artificial}) and study the resulting network in order to obtain new insights on how the brain might be organized to produce this type of memory. This methodology is similar to the one used in the field of Evolutionary Robotics(ER) where robotic controllers are evolved to study how a particular cognitive task could be accomplished by an artificial neural network (\cite{floreano10evolutionrobots}). ER has been successfully applied to the study of the evolutionary conditions for the emergence of communication in communities (\cite{floreano07evolcom}), and to the study of learning behavior in cognitive science (\cite{tuci03ecological}). The most interesting aspect of this methodology is that it reduces the influence of the researcher on the design of the model which can lead to the emergence of previously unexplored strategies to complete the task.
\par A common task in the study of STM is to monitor the persistent activity of a neural network after an initial stimulation. Our study follows a similar methodology. Our neural model consists in a spiking neural network with Izhikevich neurons (\cite{izhikevich03simplespiking,izhikevitch04modelspiking}). The network receives a pattern of activation for a duration of 1s and subsequently must maintain its output neuron active for a fixed amount of time. The memory is encoded by the capacity of the network to influence other populations of neurons as a result of the initial activity pattern. The topology of the network will be decided by the evolutionary process, but the synaptic strengths will remain fixed during the experiment. The question we explore is the type of topology that evolved, but also if a yet unknown strategy to implement STM appeared.
\par The paper is organized as follows. Section \ref{sec:methods} introduces the methods used, i.e. the details of the task with the description of the evolutionary procedure and the chosen type of spiking neurons. Section \ref{sec:results} presents our results which are discussed in section \ref{sec:discussion}. Finally the conclusion is given in section \ref{sec:conclusion}.

\section{\label{sec:methods}Methods}
The spiking neural model we rely on is the Izhikevich neuron which became a standard in neural modeling for its computational lightness combined with its relative accuracy in reproducing the electrical profile of real neurons. This model relies on a system of two differential equations:
\begin{eqnarray}
	v'&=&0.04v^{2} + 5v + 140 - u + I\\
	u'&=&a(bv-u)
\end{eqnarray}

where $v$ is the membrane potential of the neuron, $u$ its membrane recovery variable, and $I$ the weighted sum of the synaptic currents from pre-synaptic neurons. $a$ and $b$ are parameters tuning the regime of the neuron and decided experimentally. The membrane potential describes the accumulation of energy received from other neurons, while the membrane recovery implements a negative feedback pushing the membrane potential toward its resting state. When the membrane potential reaches a threshold of 30mV, a spike is emitted and transmitted to post-synaptic neurons connected to it, and the parameters are reset to $v=c$ and $u=u+d$,
where $c$ and $d$ are parameters describing the regime of the neuron. $a$, $b$, $c$ and $d$ have been provided by Izhikevich for different types of neurons. In our case, we chose to use the regular spiking model with parameters $a=0.02$, $b=0.2$, $c=-65mV$ and $d=6$. The initial state of a neuron is $v=-65$ and $u=b*v$. The synaptic currents, $I$, are computed by summing the weighted contribution from all the pre-synaptic neurons, as in the following equation
\begin{eqnarray}
I_{j} &=& \sum_{i \in S} w_{ij}*\delta{}(v_{i})\\
\delta{}(v_{i}) &=& \left\{
	\begin{array}{rl}
		0\text{mV } & \text{if } v_{i} < \text{threshold}\\
		30\text{mV } & \text{otherwise}
	\end{array} \right.
\end{eqnarray}
where $S$ is the set containing all the pre-synaptic neurons of $j$, and $w_{ij}$ is the strength of the synapse connecting $i$ to $j$. Synaptic strength is always in the range $[-1;1]$, and the inhibitory or excitatory nature of a synapse is determined by its pre-synaptic neuron. All synapses have a transmission delay of 1ms. The external input can also come from an external stimulation, such as the one we use to stimulate the network during the first second of the experiment.

\par
\begin{figure}[htbp]
\begin{center}
\includegraphics[width=5.0cm]{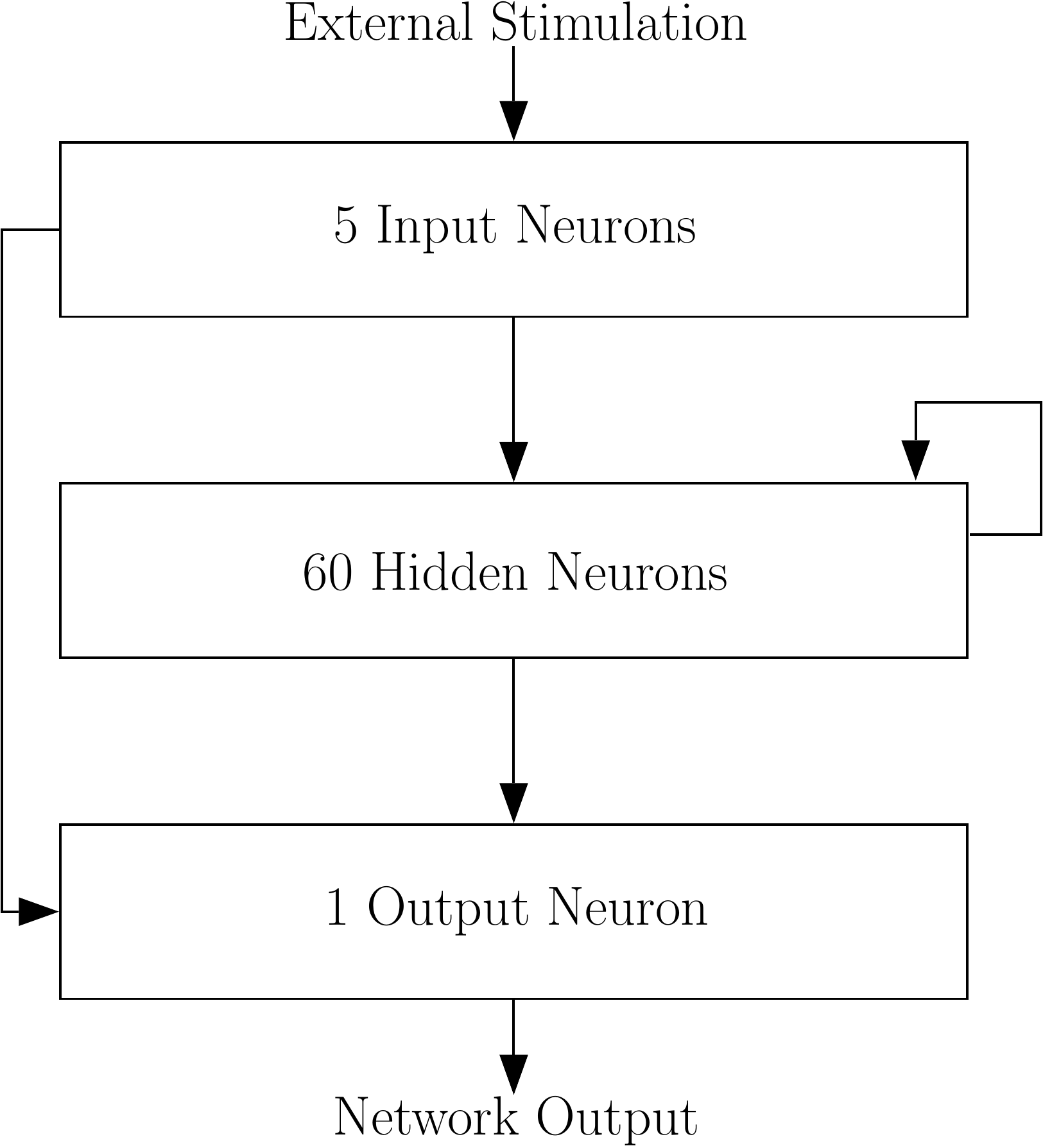}
\caption{\label{fig:nntopology}Topology of the SNN. The network is composed of five inputs neurons, 60 hidden neurons interconnected without self-recurrent connections, and one output neuron used to evaluate the activity of the network.}
\end{center}
\end{figure}

 The prototype for the networks used in the following experiments is composed of five inputs, 60 hidden neurons and one output. As shown in figure \ref{fig:nntopology}, the input neurons are connected to all hidden and output nodes, while the hidden nodes are connected to all output nodes. There is no self-recurrent connection, or feedback from output nodes to input and hidden nodes, nor from hidden nodes to input nodes. The five inputs all receive the same stimulation. The rationale for using five input nodes is to increase the capacity of the initial signal to stimulate the network. Indeed, one input node would not be sufficient to create enough spikes in the network to create the dynamics we seek. The number of five was chosen experimentally in order to amplify the effect of a single input. 
\par Every hidden node is connected to all hidden nodes outside itself. There is no restriction on the connectivity between excitatory and inhibitory neurons. The strength of the synapses, and the excitatory or inhibitory nature of each neuron, is tuned by a GA. The number of neurons has been chosen experimentally and is relatively small to facilitate the evolution. Every individual is described by 3965 genes, each represented by a real value within $[0;1]$, which required thousands of generations before obtaining an adequate configuration to solve the task, as we will show later in the result section. Among these genes, 60 genes determine the excitatory or inhibitory nature of the hidden neurons (input and output neurons are excitatory). If the value of the gene is lower than 0.5, the neuron becomes inhibitory. Otherwise, it is excitatory. The remaining 3905 genes encode the synaptic strengths. Based on the nature of the pre-synaptic neuron, the value of the gene will become negative (inhibitory) or remain positive (excitatory) before being assigned to a synapse.

\par The task required the network to maintain spike activity on the output neuron for a fixed amount of time. In this study, we evolved 10 separate networks(i.e. 10 experiments) capable of solving the task for a duration of the persistent activity ranging from 2s to 11s by increments of 1s. The durations have chosen to fit within the timeframe observed for STM in the brain. A network solving the task for a specific duration cannot solve the task for another. One run of an experiment unfolds as follows:
\begin{enumerate}
\item 1s of stimulation during which the network receives a constant input.
\item 2s to 11s of self-sustained activity depending on the network being evolved. The self-sustained activity is measured by the activity of the output neuron.
\item No spike activity on the output should be reported for at least 4s after the self-sustained phase is over.
\end{enumerate}

\par At the end of the run, the fitness of the network is computed using the following equations
\begin{eqnarray*}
\text{fitness } & = & e^{\left(-\frac{(x-s)^{2}}{2*\sigma(s,x)^{2}}\right)} \\[5pt]
\text{where } \sigma(s,x) & = &
	\begin{cases}
		0.35*s & \text{if } x \leq s\\
		10^{3} & \text{if } x > s\\
	\end{cases}
\end{eqnarray*}
where $x$ is the time of the last spike emitted by the output neuron and $s$ the time when the output is expected to stop spiking. The two equations compute how well a SNN performs with respect to the task we would like to evolve. These equations are Gaussian functions: the first increases the fitness of a network if the output neuron stops its activity around the target stopping time, and the second reduces it quickly if it passes this deadline. In our experiments, we use a simulation time step of 0.1ms which gives us $s=(1+2)/0.0001$ for a 2s period of self-sustained activity before forgetting. Time in this simulation is realistic as Izhikevich's equations are modeled following a 1ms resolution and integrated using the Euler method.
\par With the chosen topology of the SNN, and this experimental setup, we consider that there is no restriction on the topology that could potentially be evolved. The connection between the input and the output neurons is only effective during the first second of the experiment, during which the input neurons are stimulated. During the self-sustaining period, no stimulation transits between the input and output neurons. The complete dynamics resides within the hidden neurons whose topology is decided by the evolution only. We will then focus our analyses on the hidden neurons only.
\par The 10 networks presented in this paper are tuned using an evolutionary algorithm. They are subsequently tested for the task. This effectively creates two different timescales for our experiments: the generational timescale and the lifetime of the network. The generational timescale is the number of generations the evolutionary algorithm has produced. The lifetime of the network consists in the seconds of an experiment during which the fitness is computed. These two timescales are separate and have no influence on each other. The evolutionary algorithm is used to select networks capable of solving the experiment. It is not different from generating random networks and selecting the best to experiment on. The GA simply offers the advantage of producing networks that could be evolved in nature and a better convergence speed toward a solution. During the lifetime of the evolved individuals, there is no synaptic change effected on the weights. This methodology differs from machine learning methods in which the individual is adapting to its environment during its lifetime. In the rest of the paper, when we talk about plasticity, we refer only to the process occurring during the lifetime of the individual, not the tuning of the weights by the GA.

\section{\label{sec:results}Results}
\subsection{Evolution}
Using this methodology, we evolved networks capable of sustaining their activity for a specified amount of time, and extinguishing their activity past that time. Using 66 neurons, we evolved 10 networks, each capable of persistent activity for a specific amount of time ranging from 2 to 11 seconds.
\par
\begin{figure}[htbp]
\begin{center}
\includegraphics[width=10.0cm]{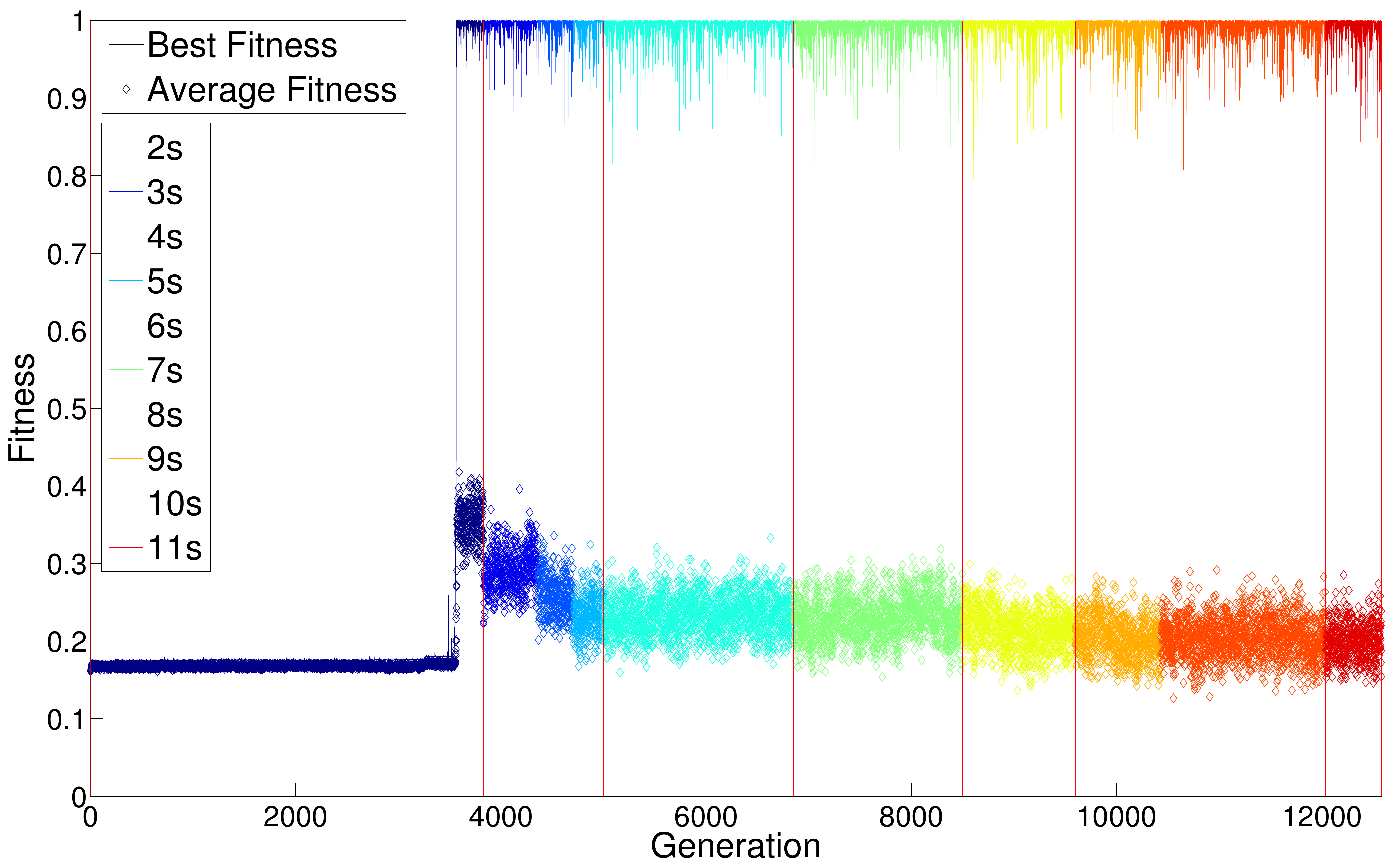}
\caption{\label{fig:evolution}Evolution of the fitness for all the evolved networks. The dotted line shows the average fitness of the population, while the continuous one the best fitness at every generation. The red vertical lines represent where the fitness switches from one duration to the next and the colors indicate which network is being evolved at each generation.}
\end{center}
\end{figure}

 Figure \ref{fig:evolution} shows the evolution of the fitness for all the networks starting from 2s up to 11s. In order to accelerate the evolution, we used continuous evolution. That is we first evolved a 2s network, and changed the fitness function to evolve a 3s network after obtaining a successful individual. We repeated this process until obtaining a 11s network. In figure \ref{fig:evolution}, the red lines separate the moments where the fitness is changed to evolve a longer duration, as is the case for the line located at generation 3829 where the fitness switched from the evolution of a network self-sustaining for 2s to one whose duration is 3s. One interesting aspect of this graph is the absence of drop in fitness at the time of the switch. A change is visible in the average fitness, but there is no clear difference for the fitness of the best individual. This implies that an individual capable of solving the next duration was already present in the population every time. This can be explained by the diversity of the population combined with the fitness function granting fitness even to individuals who are not perfect in their stopping time. However, if we look at the initial generations, where fitness remained flat for 3564 generations before jumping to the maximum fitness, we can also say that the solutions are most likely to be located within a narrow section of the evolutionary landscape. 

\par The choices of starting the evolution with a 2s network, and using continuous evolution up to 11s, are due to the fact that our fitness function facilitates the evolution of networks with a short self-sustaining period when started from a random population. The fitness function is a mapping from the stopping time of the network to a Gaussian function: networks very close to the peak of the bell curve receive a high fitness, while those on the tail receive a much lower one. As the tail is relatively flat around the peak, networks falling into that region will not see their fitness improve much even if they get much closer to the target duration. On the contrary, small changes in duration in the vicinity of the peak have huge impact on the fitness obtained. As a result of this effect, it is easier to evolve a 2s network than a 5s one because the tail before the peak in the 2s evolution is shorter than in the 5s evolution. What makes this continuous evolution successful is that the networks that do not reach the target duration, but are close enough to receive a fitness slightly higher than if they were on the tail region of the fitness curve will be promoted and maintained by the evolution. This explains why an evolution for a 2s network will most likely have a 3s network in its population, but also why it is highly unlikely to have a 5s one.
\par To estimate how likely it is to obtain a random network capable of solving a 2s task, we generated 100.000 networks and recorded the time of their last spike. Those networks showed no, or very limited, self-sustainability. We found 118 networks capable of self-sustained activity for the whole 7s composing a complete experiment. These were capable of persistent activity but not of forgetting. In-between the two extremities, only three networks self-sustained and stopped their activity past the 2s mark. None of them provided a stopping time around 3s. This demonstrates that the networks we evolved are not common within the evolutionary landscape, and are most likely not the result of a random activation of the neurons. It also shed a light on the difficulty of forgetting as we obtained many more networks capable of only self-sustaining their activity compared to the amount of networks that could self-sustain and later on become silent.

\subsection{Study of the 2s network}
\begin{figure}[htbp]
\begin{center}
\includegraphics[width=10.0cm]{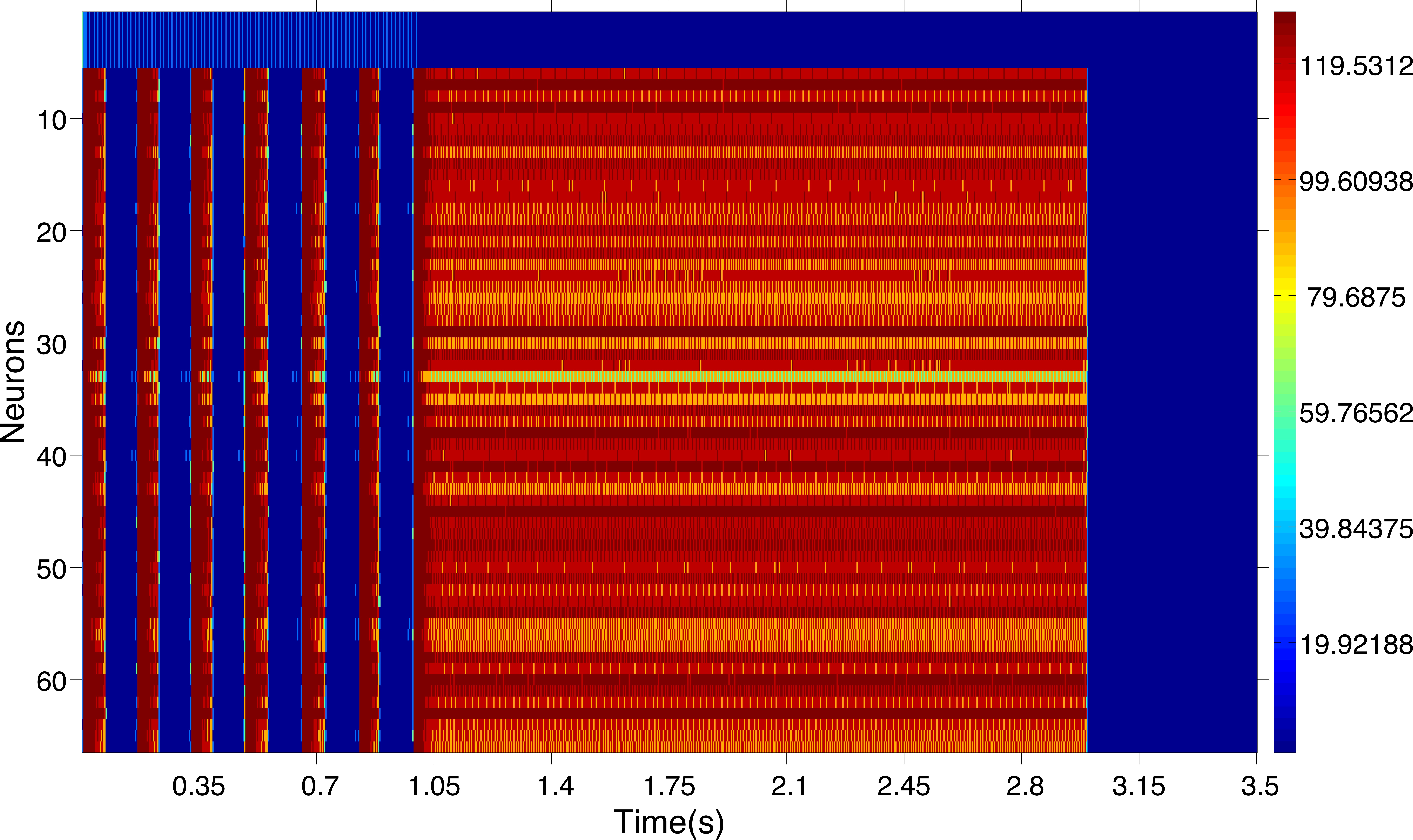}
\caption{\label{fig:rasterplot} Activity of a successfully evolved network. The x-axis is the time in seconds, and the y-axis represents the neurons. Each colored dot represents the number of spikes occurring during a fixed time window of 3.5ms, the color varying from blue to red for low to high spike counts respectively.}
\end{center}
\end{figure}

\begin{figure}[htbp]
\begin{center}
\includegraphics[width=10.0cm]{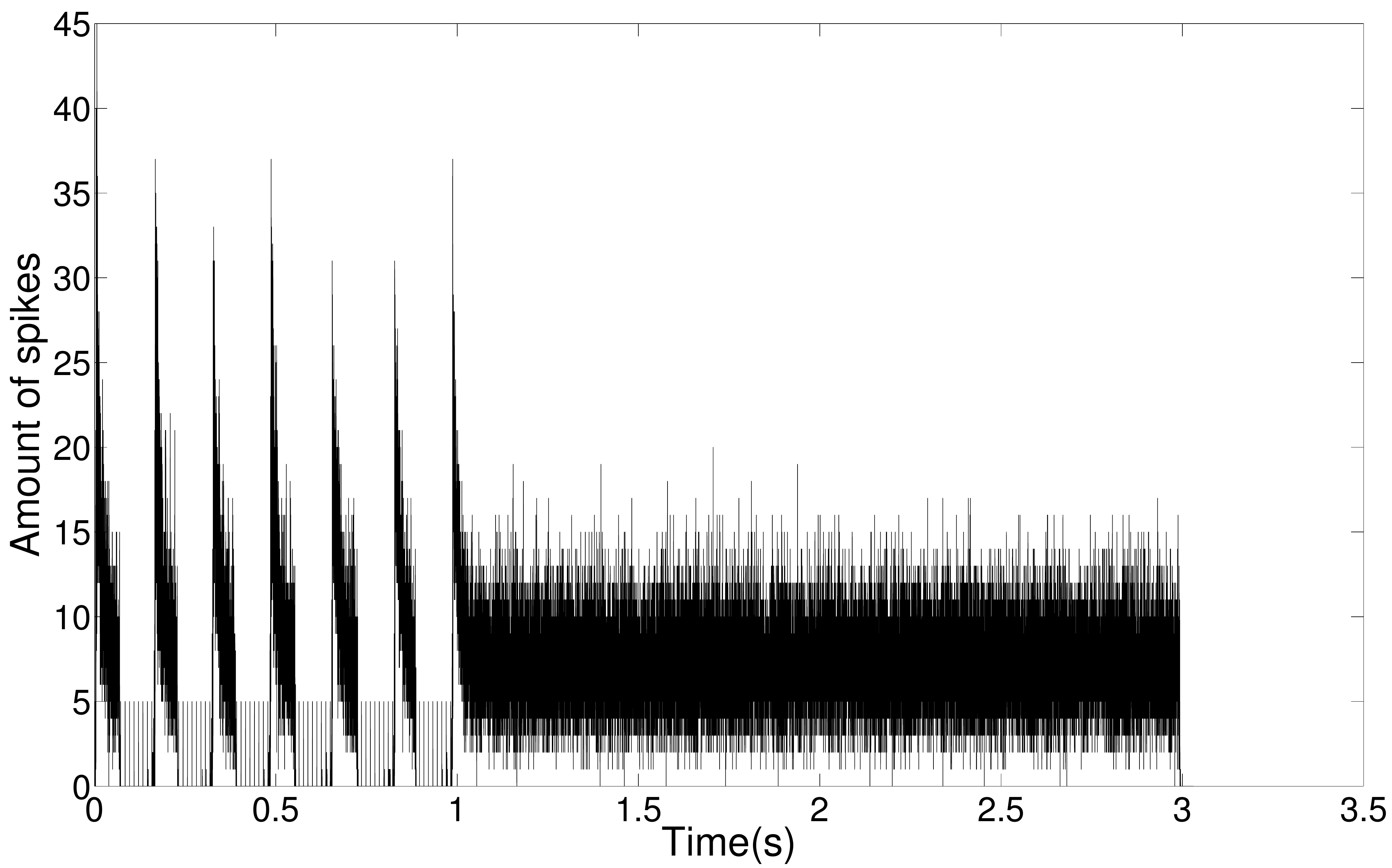}
\caption{\label{fig:spikecount} Activity of a successfully evolved 2s network shown through the quantity of spikes at every timestep. The x-axis is the time in second, and the y-axis represents the number of spikes recorded.}
\end{center}
\end{figure}

To illustrate the activity of the networks, we will focus on the network exhibiting 2s of persistent activity. The other networks show similar patterns of activation so we are confident that any conclusion drawn from this network can be transferred to networks with a longer self-sustained period. 
Figure \ref{fig:rasterplot} presents the activity of the network during the first 3.5 seconds of the experiment, while figure \ref{fig:spikecount} shows the amount of spikes at each timestep for the whole network. During the first second, the network is stimulated by a regular input pattern. This results in the network showing periods of low activity separated by high amounts of spikes for that period. At every instant, between 0 and 45 neurons spike. From 1s to 3s, the input neurons remain silent and the network must self-sustain its own activity. During this period we can see that all the neurons spike frequently. Figure \ref{fig:spikecount} shows that the maximum number of neurons spiking is lower than during the initialization phase, and that, at every instant, between 0 and 21 neurons spike. From timestep 29952 (2.9952s), the network becomes silent and no spike can be observed for the rest of the trial. It is interesting to note that no progressive decay of the neuronal activity is observed before the network becomes silent.

\par
\begin{figure}[htbp]
\begin{center}
\begin{tabular}[]{cr cr}
Stimulated Activity && Self-Sustained Activity\\
\includegraphics[width=5.5cm]{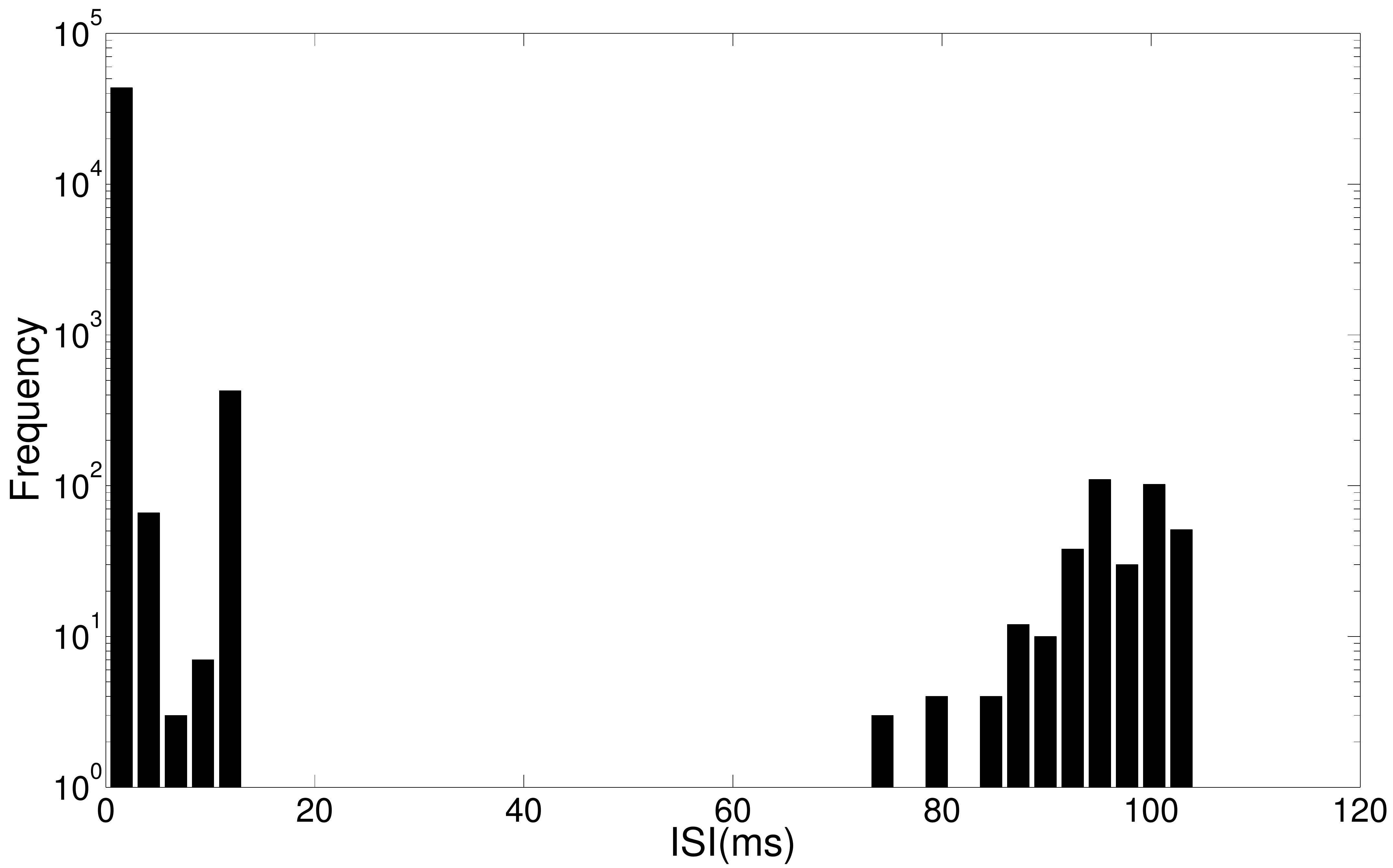} && \includegraphics[width=5.5cm]{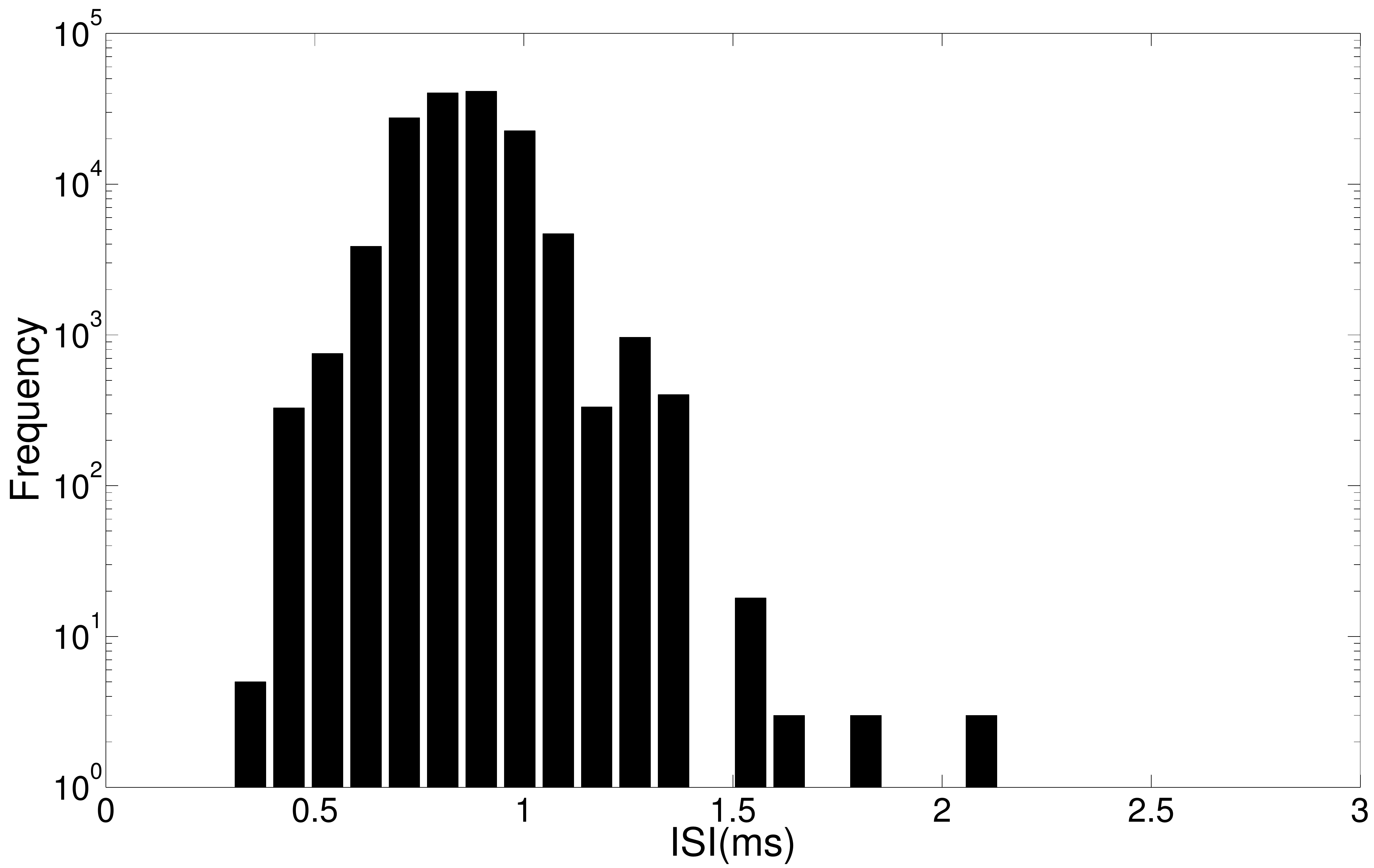}
\end{tabular}
\caption{\label{fig:isidistribution_neurons}ISI distribution of a 2s network. The graph on the left shows the distribution during the stimulation period (0s to 1s), while the graph on the right shows the distribution during the self-sustained period (1s to 3s).}
\end{center}
\end{figure}

\begin{figure}[htbp]
\begin{center}
\begin{tabular}[]{cr cr}
Stimulated Activity && Self-Sustained Activity\\
\includegraphics[width=5.5cm]{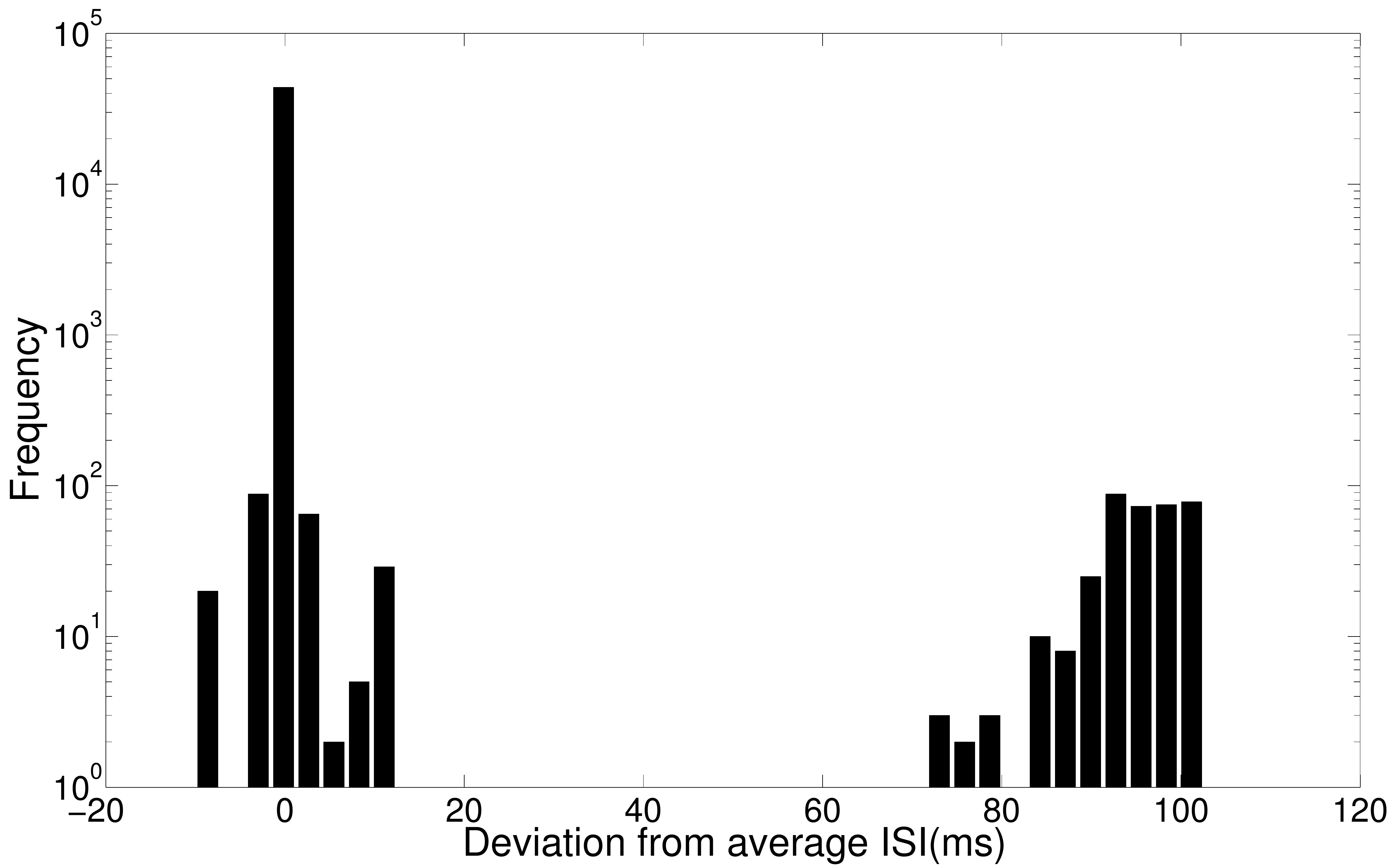} && \includegraphics[width=5.5cm]{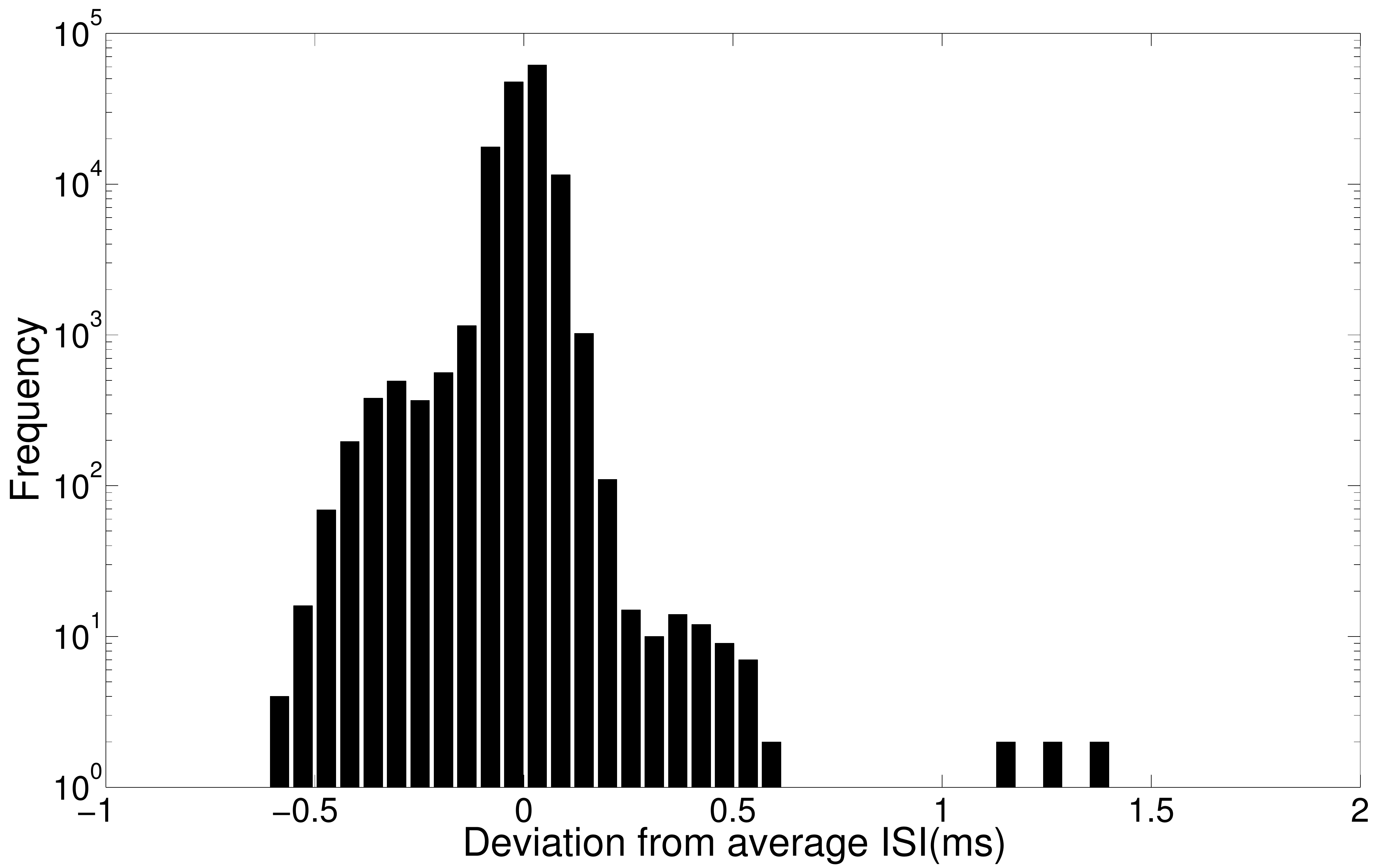}
\end{tabular}
\caption{\label{fig:isi_neurons_deviation}Distribution of the deviation of the ISI from the average ISI of every neuron for a 2s network. The graph on the left shows the deviation during the stimulation period (0s to 1s), while the graph on the right shows the deviation during the self-sustained period (1s to 3s).}
\end{center}
\end{figure}

 The analysis of the interspike intervals (ISI) shows that, during the self-sustaining period, the neurons spike very regularly. Figure \ref{fig:isidistribution_neurons} shows the distribution of the ISI of the network during the stimulated period, and the self-sustained period. Figure \ref{fig:isi_neurons_deviation} shows the histogram of the deviation from the average ISI of each individual neuron, which is computed by subtracting the mean of the ISI from the standard deviation. The distribution shows that most of the neurons spike between 0.8 and 1.2ms, with a small jitter in the range of  [-0.5;0.5] ms during the self-sustained period, as seen in figure \ref{fig:isi_neurons_deviation}. During the stimulated period, the ISI varies much more. As the activity plot showed, there are moments of intense activity followed by moments of silence during this period. We can see this tendency in the ISI distribution with most of the ISI very close to zero while some range as high as 100ms.
\par The regularity of the neurons during the self-sustained period is to be noted. It shows that the neurons are not exactly periodic, but still remain very precise in their spiking time. The rate-normalized coefficient of variation (CV), computed by dividing the standard deviation of the ISIs by the mean, is close to zero for all neurons during the self-sustained period, which further indicates that the spike trains that they exhibit are highly regular (\cite{softky93highlyirregular}). This makes it difficult to quantify how the network achieves forgetting because of the lack of periodicity, which prevents the neurons from synchronising themselves. Indeed, we clustered the neurons using their activation times and found that it was not possible to create groups of neurons according to their spike time that would remain unchanged during the duration of the experiments. There does not seem to be any strong correlation between the activation times of the neurons. There is also no difference in ISI between the neurons of the two groups, i.e. there is no correlation between the ISI of a neuron and its participation in one of the two groups.

\par
\begin{figure}[htbp]
\begin{center}
\includegraphics[width=10.0cm]{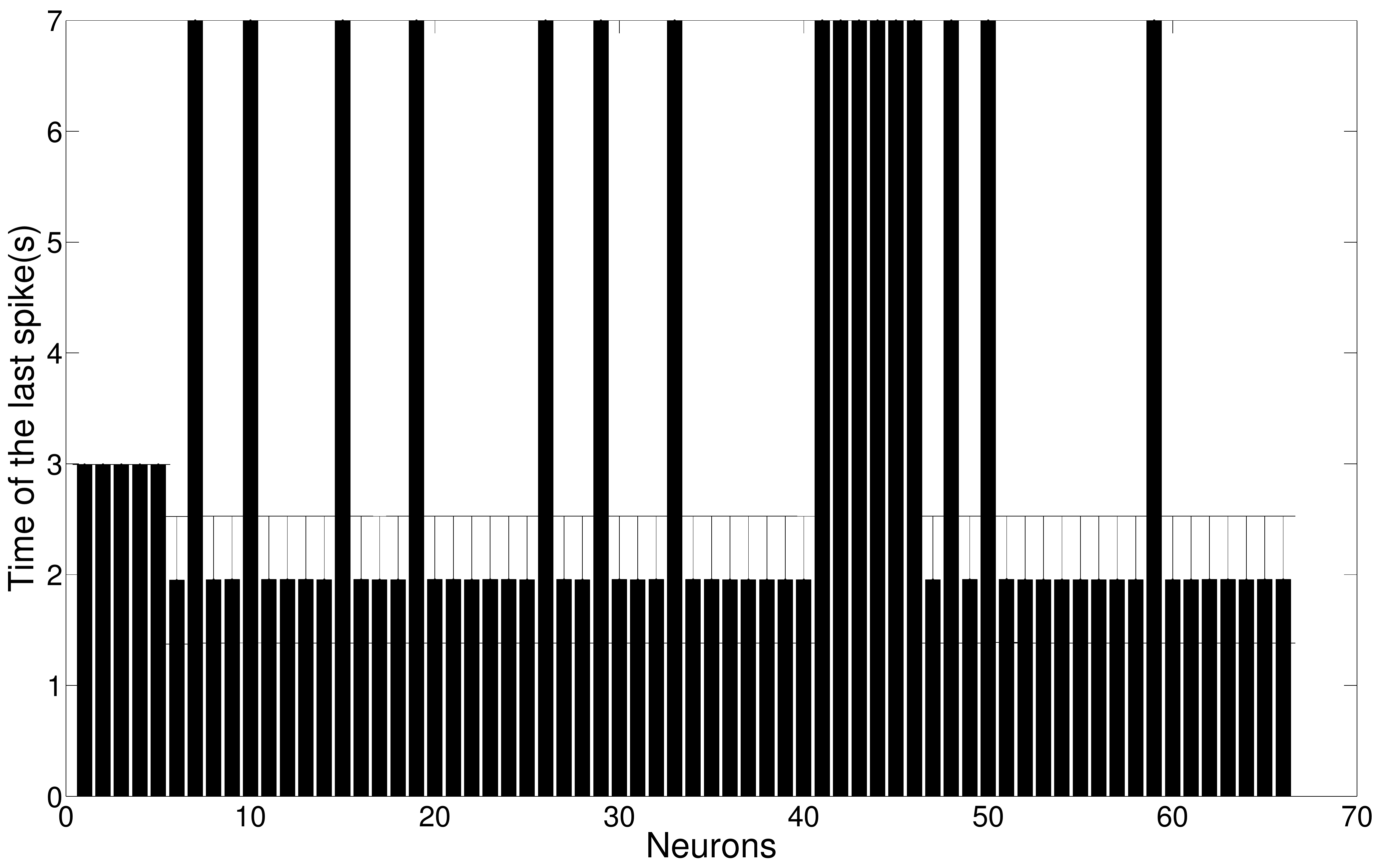}
\caption{\label{fig:pruning_summary_active} Summary of the pruning experiments. Each bar represents the average time of the last spike emitted by the output neuron when that specific neuron is pruned. The average is obtained over different pruning time varying from 1s to 2.9s with a step of 0.1s. The error bars represent the standard deviation.}
\end{center}
\end{figure}

 To understand how each neuron influences the behavior of the network, we conducted some pruning experiments. In these experiments, we deactivate each neuron individually and record the time of the last spike emitted by the output neuron, as well as compute the fitness of the network. As the impact of a neuron might vary over time, we deactivate each neuron at different intervals. More precisely, we deactivate a neuron every 0.1s during the self-sustained period lasting between 1s and 3s. Once a neuron has been deactivated, it does not emit a spike for the remainder of the trial. Figure \ref{fig:pruning_summary_active} shows the summary of the experiments for pruning times within the self-sustained period. We can see that input neurons have no impact on the stopping time as they have no activity during this period. Among the other neurons, we can clearly distinguish two groups. The first is composed of 44 neurons which have an average stopping time located around 1.9s, with a small but visible standard deviation. The second possesses 16 neurons which have an average stopping time of 7s with no standard deviation, 7s being the duration of the trial. When a neuron of the first group is deactivated, i.e. it is prevented from emitting any spike, the activity in the network stops immediately. On the other hand, if a neuron of the second group is deactivated, the network remains active and does not become silent at the expected time. The network becomes purely self-sustained. This lead us to believe that the neural network is composed of two functional groups of neurons: one group maintains the activity during the self-sustained period, while the other stops the activity at the right time.
\par To confirm this hypothesis, we deactivated all the neurons in each group and recorded the stopping time. The result of this test showed similar results: even if all the neurons in a group are deactivated, the behavior of the network is the same as when only one neuron of the group was deactivated. This clearly shows that these 2 functional networks function separately. The neurons from each group are not being shared with the other group. If that were the case, the deactivation of one group would lead to a different behavior than what was initially observed with the deactivation of single neurons. For instance, we would expect that a neuron from the stopping group which also belongs to the self-sustaining group would prevent the network from self-sustaining if it was deactivated. This was not the case in this test. As such, we can say that the two functional networks do not share neurons.

\par
\begin{figure}[htbp]
\begin{center}
\includegraphics[width=10.0cm]{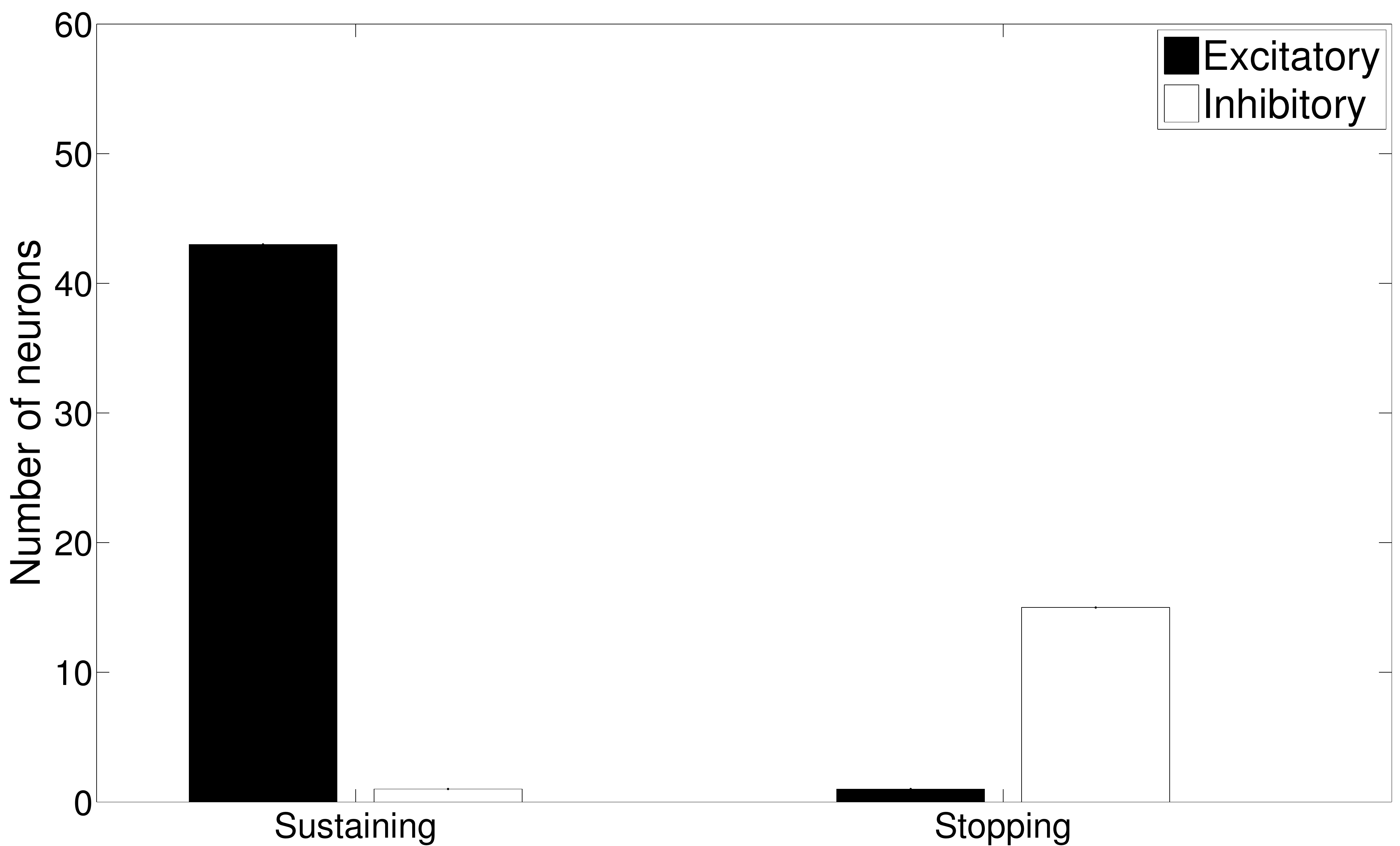}
\caption{\label{fig:excitinhib} Composition of the neuronal groups, i.e. sustaining or stopping neurons, for all durations. Each bar represents the average number of neurons in each category. The standard deviation is zero for all categories.}
\end{center}
\end{figure}

 The same analysis has been applied to the other networks, from 3s to 11s, and the same two groups have been found. Further, the composition of the groups remained the same: the same neurons remained in the same group. With regard to the excitatory to inhibitory ratio, figure \ref{fig:excitinhib} shows that there is no variation between any of the evolved networks: the stopping group is composed of all inhibitory neurons with one excitatory one, while the sustaining group possesses all excitatory neurons but one. This means that the difference between the networks lies only in the evolved synaptic strengths. But this also means that the two groups that were evolved are an evolutionarily robust strategy to solve this task.

\par
\begin{figure}[htbp]
\begin{center}
\includegraphics[width=11.0cm]{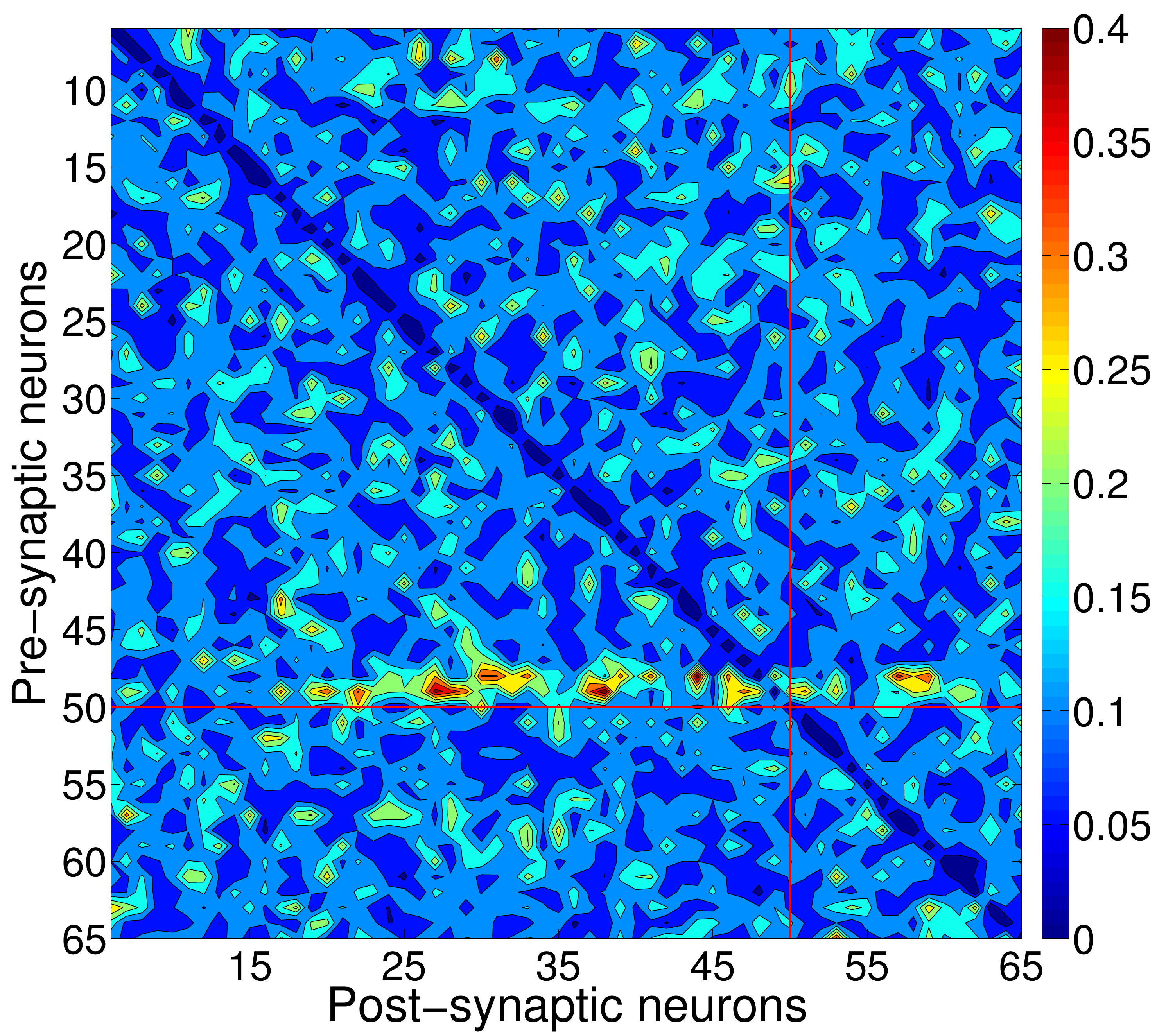}
\caption{\label{fig:weightchanges}Synaptic strength changes between networks from 2s to 11s. Changes are represented by the standard deviation of the weights of the same synapse along the different durations. The x-axis is the indexes of the post-synaptic neurons and the y-axis of the pre-synaptic ones. The colors range from blue (no change) to red (strong change). The left top section of the graph shows the changes for the synapses within the sustaining group, the top right shows the ones from the sustaining to the stopping group, the bottom left from stopping to sustaining group, and the bottom right within the stopping group.}
\end{center}
\end{figure}

 The last two experiments aim at understanding what parameters influence the duration of the persistent activity. As we already saw that the nature of the neurons remains the same between the evolved networks, we focus on the synaptic strengths in these two analyses. First, we looked at the differences between the 10 evolved networks by computing the standard deviation of the strengths of individual synapses over the 10 networks. From the results, shown in figure \ref{fig:weightchanges}, we can see that many synapses have been modified by the evolution. For most synapses, the modifications are weak and do not necessarily indicate a strong influence on the duration. The strongest variations are found for the neuron at row 49. The synapses spreading out from it have gone through stronger modifications, which might indicate this neuron's importance. Nevertheless, it does not seem plausible that the difference in duration between the networks could be caused by modifying a small amount of synapses.

\par
\begin{figure}[htbp]
\begin{center}
\includegraphics[width=11.0cm]{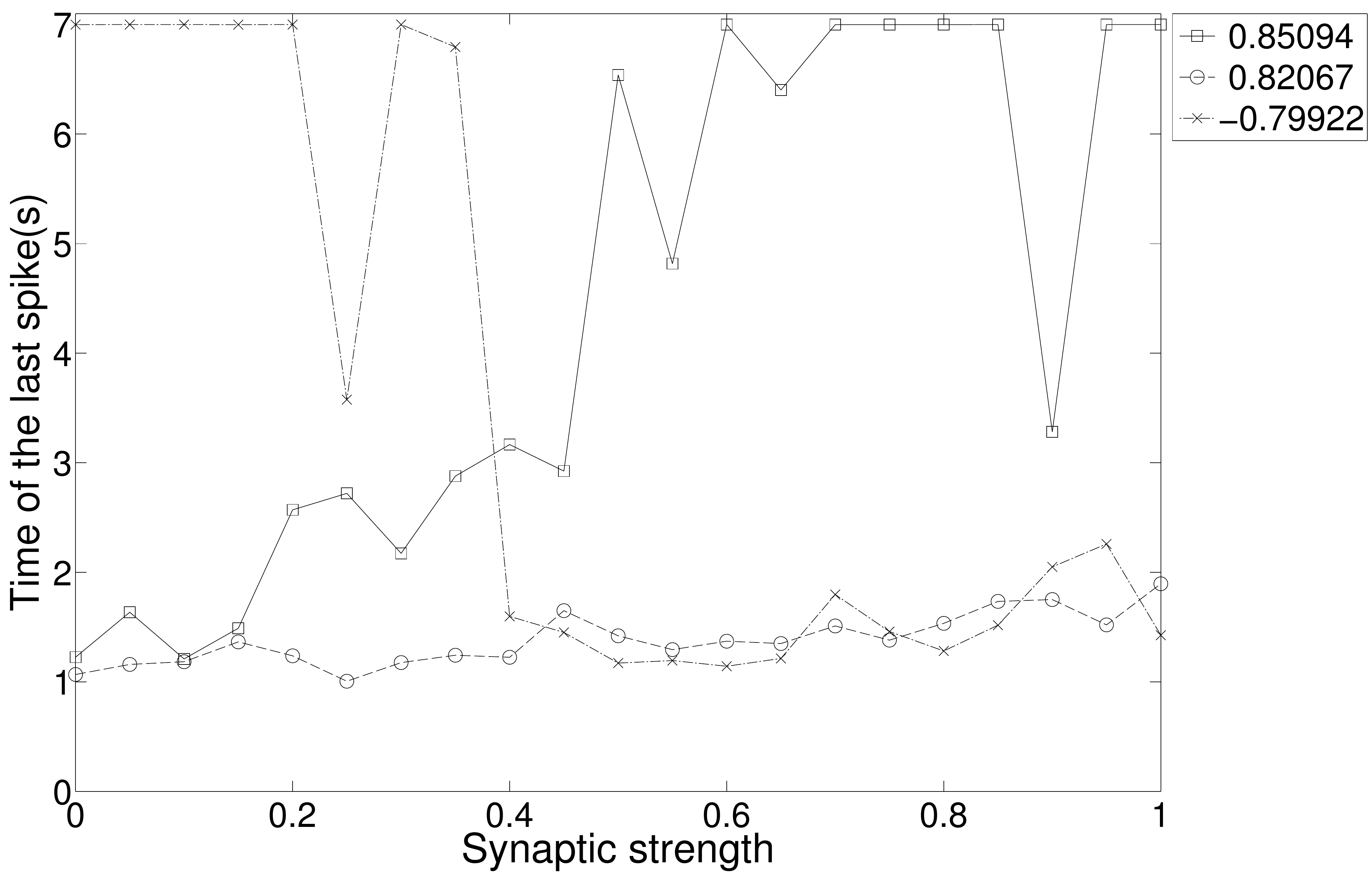}
\caption{\label{fig:weightvariation_highest}Time of the last spike for three synapses with varying strengths. The legend shows the correlation coefficients between the x and y axis. These three synapses possess the highest correlations in the network.}
\end{center}
\end{figure}

\begin{figure}[htbp]
\begin{center}
\includegraphics[width=11.0cm]{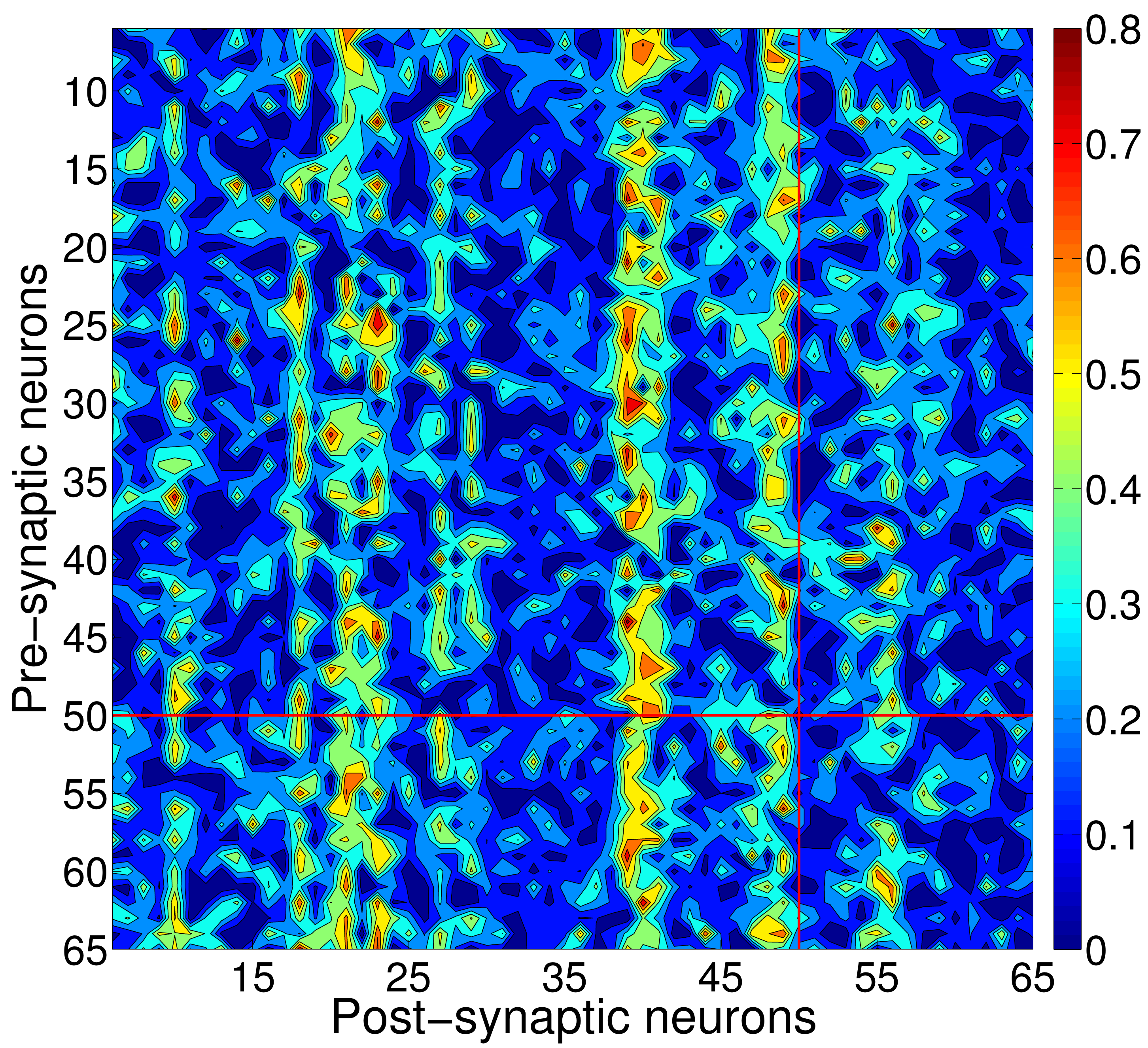}
\caption{\label{fig:weightvariation}Absolute values of the Pearson correlation coefficients between the variation of the synaptic strengths, and the duration of the persistent activity for a 2s network. The left top section of the graph shows the changes for the synapses within the sustaining group, the top right shows the ones from the sustaining to the stopping group, the bottom left from stopping to sustaining group, and the bottom right within the stopping group.}
\end{center}
\end{figure}

 To confirm this impression, we conducted another experiment on the 2s network. We varied individually the strength of every synapse, from 0.0 to 1.0 with increments of 0.05, and, for each variation, tested the network and recorded the time of the last spike emitted by the output neuron. We then computed the Pearson correlation coefficient between all the variations of a  synapse and all the associated recorded times to study the influence of this particular synapse on the activity of the network. Figure \ref{fig:weightvariation_highest} shows, as an example, the times of the last spike obtained for the aforementioned range of strengths for the three synapses with the highest correlations. A summary for all the synapses is shown in figure  \ref{fig:weightvariation}, where each point is the absolute value of the correlation coefficient for one synapse, computed using the following equation:
\begin{equation}
C_{xy} = \lvert\text{cor}(\{W_{xy}\}, L(\{W_{xy}\}))\rvert
\end{equation}
where $x$ and $y$ are the pre- and post-synaptic neurons, respectively, $\{W_{xy}\}$ is the set of tested strengths for the synapse connecting neuron $x$ to $y$, $L(\{W_{xy}\})$ is the set of the measured times of the last spike of the network for each variation of $W_{xy}$, and \emph{cor} is the Pearson correlation coefficient whose absolute value is stored in $C_{xy}$.
\par In figure \ref{fig:weightvariation}, we can see there are multiple vertical lines of higher correlations mostly surrounded by low correlation regions. The vertical lines indicate that the synapses reaching the same post-synaptic neuron have a high influence on the duration of the activity of the network. This does not indicate that one synapse is important, but that a particular neuron is important as any change in its input synaptic current leads to a correlated change on the behavior of the network. We do not see horizontal lines of synapses with high correlations as it would mean that all the neurons have a high influence on the duration of the network, which is unlikely. Some neurons are clearly more important than others. Among these, we find the same neuron that is highlighted in figure \ref{fig:weightchanges}, which reinforce its potential special status.

\par From these results, it seems difficult to establish how the duration is encoded within the networks. It appears that some neurons, rather than synapses, are important to determine the duration. Nevertheless, the number of modified synapses indicates that these cannot be ignored neither. Further investigations will determine how the networks stop their activity on time.

\section{\label{sec:discussion}Discussion}
The 10 evolved networks presented above all share the same properties: after stimulation, they maintain their activity for a fixed amount of time, and then go silent without external stimulation. The evolved solutions displayed three interesting aspects: the ease of evolving one network into another, the absence of decay in activity during the self-sustaining period, and the evolution of two functional networks with clear separate roles in the resolution of the task.
\par The ease of evolving from one duration to the next once a network stopping at 2s has been obtained implies that there is a small area within the fitness landscape where all the solutions to the task we presented may be found. It is quite plausible that the current topology can support longer self-sustained durations but examining this would require evolving all networks until failure, which is time consuming despite being interesting. Nevertheless, this ease of evolving other durations once one has appeared is a proof of concept that the mechanism we demonstrated could potentially be present in real neural systems. We recognise that our model is very limited compared to what can be observed in reality, but it offers a good start to study whether such non-synaptic plasticity could be observed in nature.
\par The absence of decay in spiking activity was a surprise to us. Having worked with rate-based neural networks on similar tasks, forgetting of a memory has always been due to a decay of the potential of one or multiple neurons responsible for the maintenance of the memory. With our networks, there was no obvious sign that the network would stop at any moment. The amount of spikes and their periodicity do not show any change before the network stops. We hypothesize that the state of the networks is moving along a single attractor until it becomes silent. Also, if we were to consider this result as a potential mechanism existing in the brain, it would be difficult to detect as there would be no sign of decay in activity, and could only be noticed by a sudden extinction of the neurons involved in the memory.
\par While not presented here, we tested the robustness of the networks. Our experiments concluded that changing randomly chosen synaptic weights, modifying the pattern used during the initial stimulation, or starting the experiments with neurons not in their resting state, lead to a failure of the networks to complete their task. The performance of the network varies non linearly with the noise applied to its synaptic weights, or to the stimulation. This absence of robustness was expected, as no noise was added to the simulation during the evolution. This omission facilitated the evolution of the networks we presented, but prevented the generation of robust networks. We nevertheless conducted evolutions with noise but could never obtain a maximum fitness under noisy conditions using the same number of neurons. It is possible that in order to absorb the noise, more neurons are necessary, but it could also require additional plastic mechanisms. In the future, we will focus on providing robustness to internal noise to our systems. Those experiments could provide us with interesting avenues to explore about how the brain copes with its noise, while maintaining a coherent behavior. It is also interesting to note that the capacity of the network to complete the task only when stimulated by the same pattern is a property that is most likely desired so that different patterns would not trigger the memory. If that were the case, the information retained by the memory could not be identified later on.
\par Because of the absence of robustness in the system, the presence of the two groups of functional networks, i.e. self-sustaining and stopping neurons, is the most surprising. The possibility of completely deactivating the 16 neurons composing the stopping group without compromising the self-sustainability provided by the other group is unexpected. Our initial assumption was that all the neurons were interacting following a tight dynamic. Rather, it seems that what drives the network is the interaction between the two groups. We are currently unable to quantify how the two groups communicate in order to solve the task. And if it is indeed one attractor being followed by the network, it might be extremely difficult to understand its internal dynamics. Nevertheless, the evolution of those two groups with two different roles matching the requirements of the task opens the idea that evolution might promote the breaking up of complex tasks into simpler ones at the neuronal level. Also, the fact that the stopping group is composed primarily of inhibitory neurons interacting with the mostly excitatory neurons of the self-sustaining group hints at the presence of a hierarchy in terms of control in neural systems. It would be extremely interesting if the same characteristics remained in spite of the presence of noise, as it would make it more plausible that the brain could share the same properties at the functional level.

\par Compared to other models of STM, our models have similarities and differences. Models of STM, generally referred to as models of working memory in the following literature, can be divided into synaptic and non-synaptic models. Synaptic models are based on the involvement of synaptic modifications to encode the memory either by reinforcing the synapses of a chain of neurons reacting to a specific stimuli (\cite{szatmary10working}), or by fast synaptic modifications in correlation with the activity of the neurons (\cite{sugase2008short}). Those models are further away from ours as our synapses remain static during the experimentation. Our models can be included within the category of non-synaptic models. As mentioned earlier, memory is obtained through the persistent activity of a group of neurons. Different mechanisms have been proposed to that end: the dynamics of membrane currents (\cite{marder1996memory}), the reciprocal excitation between large groups of neurons (\cite{wang2001synaptic}), reverberation within a group of neurons (\cite{GoldmanRakic1995cellular}), or through NMDA receptors mediated recurrent excitatory connections (\cite{lisman1998role, tegner02dynamical,wang2001synaptic}).
\par Our models are similar to 'bump' attractor models (\cite{wang2001synaptic, funahashi89mnemonic}). In these models, working memory is implemented through the activation of set of neurons by an initial pattern followed by a persistent activity implemented by the network going through multiple attractors, i.e. multiple network states. Our networks show a similar property as the state of the network seemingly navigates within one or multiple attractors before suddenly disappearing. One particularity of the attractor networks is that they do not withstand perturbations well. They require finely tuned synaptic strengths to maintain the attractors containing the memory. We found a similar result with our experiments when applying noise to the synapses.
\par Another aspect of our model is the strong dichotomy between excitatory and inhibitory neurons. These two types of neurons are distributed among two different functional modules within our evolved networks. This aspect is seldom found in other models of STM. Most of them do not specifically address the excitatory nature of the neurons, or the importance of the inhibitory neurons. Nevertheless, there is evidence that shows that interaction between inhibitory and excitatory neurons is important for STM, as studies of the prefrontal cortex found (\cite{GoldmanRakic1995cellular}). More recent research has proposed that the inhibitory neurons act to limit the excitatory currents in order to maintain the neurons into a sub-threshold regime where a slight increase in inputs create a spike train differing from the default activation of the network (\cite{barbieri2008can,renart06meandriven,roudi07balanced}). The excitatory-inhibitory interactions are necessary in a model to reproduce what is being observed in the prefrontal cortex. Whether our model reproduces this effect is unclear, but the regularity of the spike trains observed does not support this hypothesis.
\par With regard to the sudden extinction of the neuronal activity at the required time, other models do not display this particularity. Instead, they demonstrate a graceful decay of activity. The difference is most likely a result of their lack of requirements on the duration of the maintenance of the memory. Similar extinction has been found in neural models of temporal duration, but these rely on simpler models of spiking neurons that do not implement refractory periods, or are naturally susceptible to this type of decay (\cite{okamoto00temporalduration,bugmann98timing}). As such, it is not clear if the cause of the sudden extinction is similar to those types of models, even though our model can also be considered to be a model of temporal duration because of the fixed duration of the neural activity.

\section{\label{sec:conclusion}Conclusion and Future Work}
In this work we presented 10 spiking neural networks capable of implementing a short-term memory for a fixed duration and not relying on any form of synaptic plasticity. The memory is seen as the maintenance of the activity of the network after having been stimulated by an external source that could be another neuronal assembly. Those networks also implement forgetting through the extinction of this activity after a certain amount of time.
\par
\begin{figure}[htbp]
\begin{center}
\includegraphics[width=10.0cm]{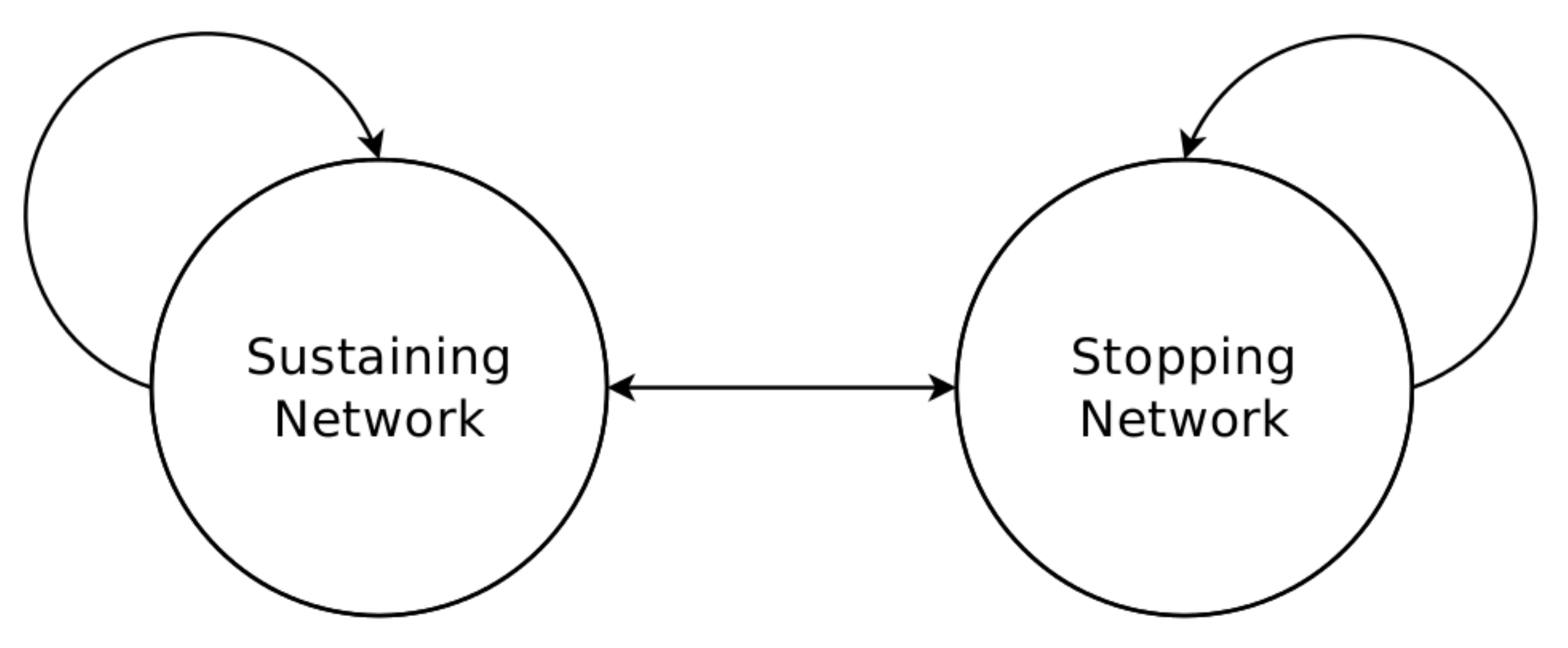}
\caption{\label{fig:network_diagram}Diagram representing the two functional networks interacting to complete the task.}
\end{center}
\end{figure}

 The evolved networks show that a short-term memory does not necessarily need to rely on the decay of the activity of its constituent neurons in order to implement forgetting. The networks in this research maintain a strong activity before becoming silent within a few milliseconds. In addition, the evolved networks showed a separation of the two sub-tasks, that is sustaining the memory and its forgetting, through evolution of two functional networks, each responsible for one sub-task. In essence, these two functional networks communicate in order to complete the task as represented in figure \ref{fig:network_diagram}. An open question from this result is what would the evolved strategy, and topology, be for other evolved tasks. As we have seen with this research, it is possible that other tasks could also lead to a separation of the neural network into modules independently implementing sub-tasks, and communicating in order to achieve the full task that is required. One interesting question would then be whether those functional networks can be transferred to other tasks sharing similar functional modules.

\par These results are the first step toward developing an understanding of how the dynamics of a group of neurons can implement a simple form of memory without requiring any structural modifications. The next step of this research will involve robustness to internal noise on the membrane potential of the neurons, but also on the pattern of the initial stimulation. Particular attention will be given to the addition of self-sustaining neurons, as preliminary results seem to indicate their presence can lead to very different strategies.
\par Another future improvement will be to implement these models inside a robotic model, and replace the artificial task used to evolved our models by an embodied one such as the navigation of a t-maze. This would show that memory based on persistent neuronal activity can exist despite the noise present in a real environment. This would provide us with a proof of concept that non-synaptic memory could have evolved in living systems existing today. 

\section*{Acknowledgements}
This work was supported by Grant-in-Aid for Scientific Research on Innovative Areas (\#24120704 "The study on the neural dynamics for understanding communication in terms of complex heterogeneous systems").

\end{document}